\documentclass[11pt,a4paper]{article}

\usepackage{cite}
\usepackage{amsmath,amssymb,amsfonts}
\usepackage{graphicx}
\usepackage{textcomp}
\usepackage[utf8]{inputenc}
\usepackage[T1]{fontenc}

\usepackage{microtype}
\usepackage{multirow}
\usepackage{booktabs}
\usepackage{array}

\usepackage{xcolor}
\usepackage{listings}
\usepackage{algorithm}
\usepackage{algpseudocode}

\usepackage{xurl}
\usepackage[hidelinks]{hyperref}
\newcommand{\code}[1]{\texttt{\detokenize{#1}}}

\lstdefinestyle{prompt}{
  basicstyle=\footnotesize\ttfamily,
  breaklines=true,
  breakatwhitespace=false,
  columns=fullflexible,
  showstringspaces=false,
  frame=none,
  keepspaces=true
}

\usepackage[margin=1in]{geometry}

\usepackage{authblk}

\begin{document}

\title{EdgeJury: Cross-Reviewed Small-Model Ensembles for Truthful Question Answering on Serverless Edge Inference}

\author[1]{Aayush Kumar\thanks{Corresponding author: akuma102@uic.edu}}
\affil[1]{University of Illinois Chicago, Chicago, IL 60607 USA}

\date{}

\maketitle

\begin{abstract}
Hallucinations hinder reliable question answering, especially in resource-constrained deployments where using frontier-scale models or retrieval pipelines may be impractical. 
We present \emph{EdgeJury}, a lightweight ensemble framework that improves truthfulness and robustness using only small instruction-tuned language models (3B--8B) suitable for serverless edge inference. 
EdgeJury orchestrates four stages: (1) parallel role-specialized generation, (2) anonymized cross-review with structured critiques and rankings, (3) chairman synthesis that integrates the strongest content while addressing flagged issues, and (4) claim-level consistency labeling based on inter-model agreement.
On TruthfulQA (MC1), EdgeJury achieves 76.2\% accuracy (95\% CI: 72.8--79.6\%), a +21.4\% relative improvement over a single 8B baseline (62.8\%), and outperforms standard baselines including self-consistency and majority voting under transparent compute accounting (total tokens and platform cost reported).
On a 200-question adversarial EdgeCases set, EdgeJury yields +48.2\% relative gains (95\% CI: 44.0--52.4\%). Manual analysis on 100 incorrect answers shows approximately 55\% reduction in factual hallucination errors versus the single-model baseline. Deployed on Cloudflare Workers AI, EdgeJury achieves 8.4~s median end-to-end latency, demonstrating that coordinated small-model ensembles can improve truthfulness on misconception-heavy QA benchmarks (TruthfulQA and EdgeCases) without external retrieval or proprietary large-model APIs.
\end{abstract}


\section{Introduction}\label{sec:intro}
Large language models have achieved strong performance across knowledge and reasoning tasks, yet \emph{truthfulness} remains a central weakness. Models may confidently assert incorrect facts or echo widely repeated misconceptions, undermining reliability in user-facing settings \cite{lin2022truthfulqa}. TruthfulQA \cite{lin2022truthfulqa} was introduced to measure this failure mode: even very large models often produce answers that resemble common human falsehoods. Later work reported \emph{inverse scaling} phenomenon on truthfulness-style evaluations, where larger models can become less truthful because they better imitate popular but incorrect text patterns \cite{mckenzie2023inverse}. Recent surveys and empirical studies consistently report that hallucinations remain a persistent failure mode across model families, especially under underspecified or adversarial prompts \cite{ji2023surveyhallucination,sahoo2024hallucinationfm}.

Despite rapid progress, most practical factuality mitigations fall into two buckets: (i) single-model sampling/aggregation (e.g., self-consistency), or (ii) retrieval-heavy pipelines that require external corpora, indexes, and extra infrastructure. These approaches are often mismatched to serverless edge settings, where the dominant constraints are orchestration simplicity, bounded calls, and predictable latency. EdgeJury targets this gap: it uses only edge-friendly SLMs, performs a single-round anonymized peer review to surface concrete failure modes, and synthesizes a final answer that integrates minority-correct reasoning rather than discarding it via voting.

Many interventions improve factuality and reasoning, but most rely on a \emph{single} model reasoning in isolation. Chain-of-thought prompting \cite{wei2022chain} and self-consistency \cite{wang2022self} reduce reasoning variance by sampling multiple trajectories and selecting a consensus answer. Reinforcement learning from human feedback (RLHF) improves instruction-following and can reduce undesirable outputs \cite{ouyang2022training}, but requires expensive annotation and typically targets a single model. Retrieval-augmented generation can improve factual grounding, but adds system complexity, latency, and dependencies that may be undesirable in edge deployments.

At the same time, deploying frontier-scale models is often impractical: large LLMs (e.g., GPT-4 \cite{openai2023gpt4}) require expensive cloud GPUs and introduce additional latency. Recent progress in \emph{small language models (SLMs)} \cite{touvron2023llama,gunasekar2023textbooks} motivates an alternative question: \emph{can a small set of edge-friendly models, orchestrated effectively, outperform a single stronger model on truth-seeking tasks under tight compute and latency budgets?}

\begin{figure*}[!t]
\centering
\includegraphics[width=0.9\textwidth]{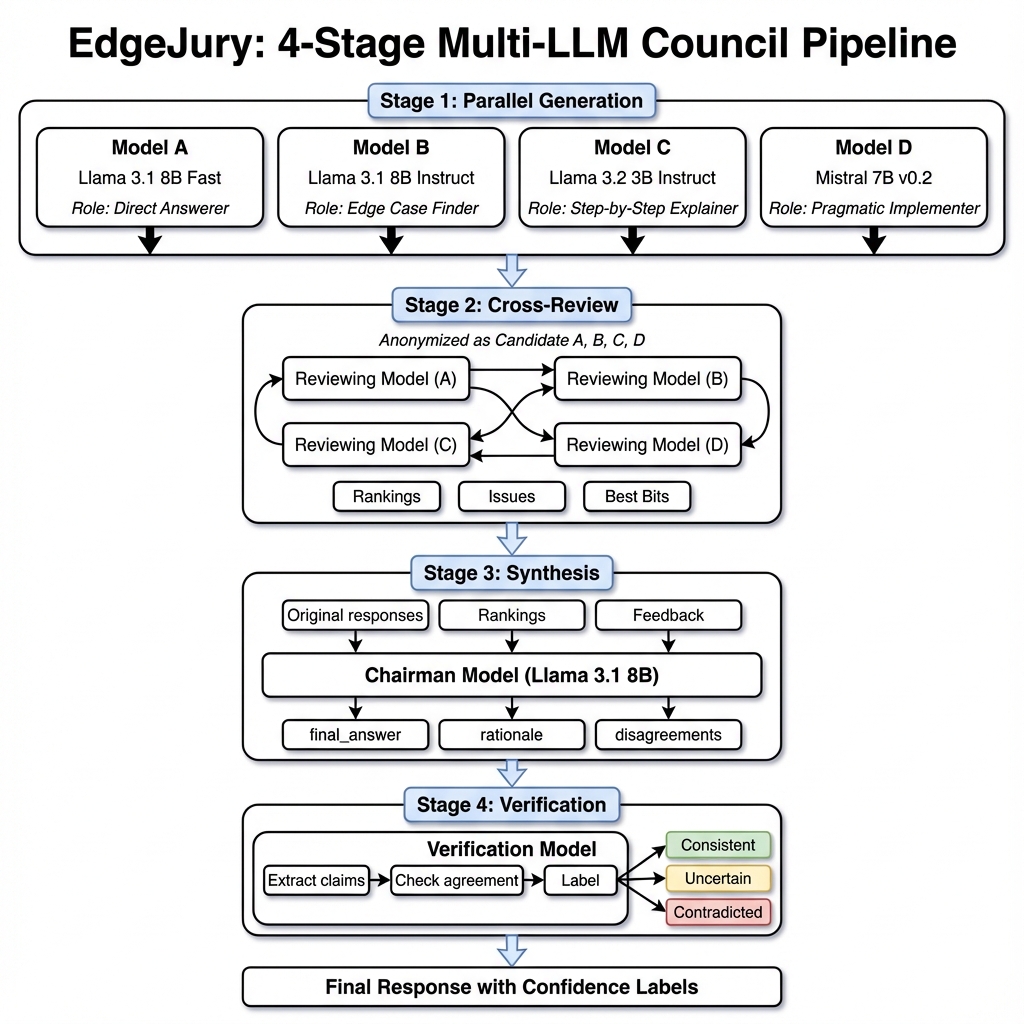}
\caption[EdgeJury four-stage pipeline]{EdgeJury four-stage ensemble pipeline. (a) Stage 1: Multiple small LLMs (3B--8B) generate role-specialized answers in parallel. (b) Stage 2: Models anonymously cross-review peers, producing structured rankings and critiques. (c) Stage 3: A chairman model synthesizes a final answer using candidate responses and review feedback. (d) Stage 4: An agreement-based verifier extracts atomic claims and labels each as consistent, uncertain, or contradicted based on inter-model agreement.}
\label{fig:architecture}
\end{figure*}

We present \textbf{EdgeJury}, an ensemble framework designed for truthful question answering with edge-deployable models. EdgeJury uses a four-stage ``AI council'' pipeline (Fig.~\ref{fig:architecture}): (1) parallel role-diverse generation, (2) anonymized cross-review that forces explicit critique and ranking, (3) chairman synthesis that integrates best components while addressing reviewer-flagged issues, and (4) claim-level consistency verification based on inter-model agreement, producing interpretable confidence tags.

Unlike multi-round debate systems that can incur high latency and many model calls, EdgeJury is intentionally \emph{single-round} for generation and review to remain practical on serverless edge infrastructure. Edge network deployment can reduce user-perceived latency by placing orchestration closer to clients and avoids maintaining dedicated GPU servers, while still enabling model coordination at practical interactive latencies. The key hypothesis is that \emph{diverse failure modes} across different small models (and roles) allow peers to catch each other's mistakes, and that a synthesis step can turn those critiques into higher-quality final answers than voting alone.

\begin{table}[!t]
\centering
\caption{System positioning. MM=multi-model; Cr=explicit critique; Syn=synthesis; NR=no external retrieval required; Edge=practical on serverless edge inference (bounded calls, predictable latency).}
\label{tab:positioning}
\footnotesize
\begin{tabular}{lccccc}
\toprule
\textbf{Method} & \textbf{MM} & \textbf{Cr} & \textbf{Syn} & \textbf{NR} & \textbf{Edge} \\
\midrule
Self-Consistency & N & N & N & Y & Y \\
Majority Vote    & Y & N & N & Y & Y \\
Debate (multi-rd)& Y & Y & Varies & Y & N \\
RAG pipelines    & N/Y & N & Y & N & N/Y \\
\textbf{EdgeJury}& \textbf{Y} & \textbf{Y} & \textbf{Y} & \textbf{Y} & \textbf{Y} \\
\bottomrule
\end{tabular}

\end{table}

\textbf{Contributions:} This paper makes the following contributions:
\begin{itemize}
\item \textbf{EdgeJury pipeline:} A practical, deployable four-stage orchestration of small LLMs that combines role specialization, anonymized cross-review, synthesis, and claim-level consistency labeling.
\item \textbf{Cross-review mechanism:} A structured, anonymized peer-review protocol producing rankings and issue lists; removing cross-review costs 7.6 absolute points on TruthfulQA MC1 (Table~\ref{tab:ablation}).
\item \textbf{Empirical results with compute accounting:} EdgeJury reaches 76.2\% on TruthfulQA MC1 vs.\ 62.8\% single-model, and outperforms self-consistency and majority vote; we report total tokens and platform cost for each method for transparency (Section~\ref{sec:budget}, Appendix~\ref{app:cost}).
\item \textbf{Reliability analysis:} Manual error analysis shows 55\% fewer hallucination errors than a single-model baseline, and Stage 4 consistency tags achieve 94.2\% precision for ``consistent'' claims on a manually checked subset (Section~\ref{sec:labelquality}).
\item \textbf{Edge deployment case study:} An end-to-end deployment on Cloudflare Workers AI with measured latency and resource usage, demonstrating feasibility without dedicated GPU servers (Section~\ref{sec:deployment}).
\end{itemize}

\textbf{Organization:} Section~\ref{sec:related} reviews related work. Section~\ref{sec:method} describes the EdgeJury pipeline. Section~\ref{sec:deployment} details deployment on Cloudflare Workers AI. Section~\ref{sec:eval} presents evaluation protocol. Section~\ref{sec:results} reports results, ablations, and analyses. Section~\ref{sec:discussion} discusses why the approach works, limitations, and future directions.

\section{Related Work}\label{sec:related}

\subsection{Truthfulness and Hallucinations}
TruthfulQA \cite{lin2022truthfulqa} formalizes the tendency of LMs to imitate human falsehoods rather than provide truthful answers. Inverse scaling results further highlight that increasing model size does not automatically yield better truthfulness \cite{mckenzie2023inverse}. RLHF can improve instruction-following and reduce some harmful behaviors \cite{ouyang2022training}, but truthfulness on adversarial misconception questions remains challenging, particularly without external grounding.

\subsection{Reasoning Prompts and Self-Consistency}
Chain-of-thought prompting \cite{wei2022chain} elicits intermediate reasoning steps and improves performance on multi-step tasks. Self-consistency \cite{wang2022self} aggregates multiple sampled reasoning paths to reduce variance and can improve accuracy without additional training. However, these methods still rely on a \emph{single} model; systematic blind spots and shared misconceptions can persist across samples.

\subsection{Ensembling and Multi-Agent Deliberation}
Ensemble methods are a classical approach to improve predictive performance through diversity \cite{dietterich2000ensemble,zhou2012ensemble}. In the LLM setting, simple majority voting and multi-sampling offer moderate gains but do not explicitly exchange critiques or integrate complementary partial solutions. AI debate proposes that adversarial argumentation can surface truth \cite{irving2018debate}. Multi-agent factuality improvements via debate-style critique have been reported \cite{du2023improving}. EdgeJury draws inspiration from these ideas but targets \emph{edge constraints}: limited rounds, small models, and an explicit synthesis stage designed to act on critiques rather than merely vote. Unlike debate systems requiring multiple rounds \cite{irving2018debate,chen2025debate}, EdgeJury uses a single-round review for low latency. 

\subsection{Self-Refinement and Cross-Checking}
Iterative self-improvement methods such as Reflexion \cite{shinn2023reflexion} and Self-Refine \cite{madaan2023self} prompt a model to critique and revise its own output. A central limitation is that a model may fail to notice its own errors. EdgeJury addresses this by using \emph{cross-model critique} and heterogeneous roles, increasing the chance that at least one reviewer detects an issue.

\subsection{LLM-as-a-Judge and Critique-Based Evaluation}
Recent work uses language models as evaluators (``judges'') to rank or critique model outputs, enabling preference-style selection without human labels \cite{zheng2023judging}. While LLM-as-a-judge is primarily used for \emph{benchmarking} \cite{guo2024survey}, EdgeJury employs critique in-the-loop as a \emph{mechanism} to improve answers under edge constraints, with anonymization and a fixed issue taxonomy to make feedback actionable.

\subsection{Post-hoc Hallucination Detection and Revision}
A common alternative direction to multi-agent deliberation is \emph{post-hoc} hallucination detection or factuality improvement via revision. SelfCheckGPT \cite{manakul2023selfcheckgpt} proposes zero-resource, black-box hallucination detection by comparing stochastic samples for consistency signals, without requiring external databases or access to model probabilities. In contrast, revision-based pipelines use external evidence to rewrite and improve factuality; RARR \cite{gao2023rarr} (Retrieve-and-Revise) iteratively retrieves supporting information and revises model outputs for better groundedness. EdgeJury is complementary: it improves truthfulness through structured cross-review and synthesis among edge-friendly models, without requiring retrieval infrastructure, while Stage~4 provides an internal agreement-based reliability signal.

\subsection{Small Models and Edge Deployment}
Recent work highlights the viability of small instruction-tuned models for practical deployments \cite{touvron2023llama,gunasekar2023textbooks} including fine-tuning of Llama-3 variants \cite{dubey2024llama}. Edge AI surveys emphasize latency, cost, and privacy benefits of pushing inference toward the edge \cite{deng2023ai}. We provide a concrete deployment and measurement study of a multi-model pipeline on serverless edge infrastructure, complementing prior work focused on model compression and general deployment considerations \cite{han2021pre}.

\section{Methodology}\label{sec:method}

\subsection{Pipeline Overview}
EdgeJury processes a user question with four sequential stages (Fig.~\ref{fig:architecture}). The design goal is to maximize \emph{useful diversity} early (Stage 1), enforce explicit critique (Stage 2), consolidate into a single high-quality answer (Stage 3), and expose uncertainty transparently (Stage 4), while keeping compute bounded. In our implementation, EdgeJury uses a constant 10 model calls per query (4 generation + 4 cross-review + 1 synthesis + 1 verification), with Stage 1 and Stage 2 parallelizable.

\subsection{Notation}
Let $\mathcal{M}=\{m_1,\dots,m_4\}$ denote the four base models (or model instances) used in Stage~1 with role prompts $\mathcal{R}=\{r_1,\dots,r_4\}$. Let $a_i$ be the answer produced by $(m_i,r_i)$ in Stage~1. In Stage~2, each reviewer produces a structured review object $\rho_j$ over anonymized candidates. Stage~3 produces the final answer $\hat{a}$. Stage~4 extracts a set of atomic claims $\mathcal{C}(\hat{a})$ and outputs per-claim agreement labels.

\begin{algorithm}[!t]
\caption{EdgeJury Inference Pipeline (per query)}
\label{alg:edgejury}
\small
\begin{algorithmic}[1]
\Require Question $q$, models $\mathcal{M}$ with roles $\mathcal{R}$
\Ensure Final answer $\hat{a}$ and claim labels $L(\mathcal{C})$
\State \textbf{Stage 1:} In parallel, generate candidates $a_i \leftarrow m_i(q; r_i)$ for $i=1..4$
\State \textbf{Stage 2:} For each reviewer $m_j$, produce review $\rho_j \leftarrow m_j(\{a_i\}_{i=1}^4)$ with anonymized IDs
\State Aggregate rankings via Borda count; aggregate issue flags via de-duplication
\State \textbf{Stage 3:} Synthesize $\hat{a}_{json} \leftarrow m_{\text{chair}}(q, \{a_i\}, \text{Agg}(\{\rho_j\}))$ 
\State \textbf{Output:} if $q$ is MC, return only $\hat{a}\leftarrow$ \texttt{choice} from $\hat{a}_{\mathrm{json}}$; else return $\hat{a}\leftarrow$ \texttt{final\_answer} from $\hat{a}_{\mathrm{json}}$
\State \textbf{Stage 4:} Extract atomic claims $\mathcal{C}\leftarrow m_{\text{ver}}(\hat{a},\{a_i\})$ and compute labels $L(\mathcal{C})$
\State \Return $\hat{a}, L(\mathcal{C})$
\end{algorithmic}
\end{algorithm}

\subsection{Stage 1: Role-Specialized Parallel Generation}
EdgeJury uses four role prompts to elicit complementary outputs:
\begin{enumerate}
\item \textit{Direct Answerer}: concise, accurate answer with explicit assumptions.
\item \textit{Edge Case Finder}: identifies exceptions, risks, and hidden assumptions.
\item \textit{Step-by-Step Explainer}: structured reasoning and derivations when needed.
\item \textit{Pragmatic Implementer}: actionable advice, examples, or procedures.
\end{enumerate}
Role specialization reduces correlation between outputs and increases coverage. In our implementation, we use two instances of LLaMA-3.1-8B (different role prompts), one LLaMA-3.2-3B, and one Mistral-7B \cite{jiang2023mistral} (Section~\ref{sec:deployment}). LLaMA-3.2-3B was chosen for its efficiency on edge infrastructure. Using multiple model families (LLaMA and Mistral) plus role-specialized prompts reduces correlated failure modes compared to sampling a single model repeatedly.

\subsection{Stage 2: Anonymized Cross-Review}
Each model reviews peer answers anonymized as Candidate A/B/C/D to reduce positional or identity bias. Review outputs are constrained to a structured JSON schema containing: (i) per-candidate ratings (accuracy/insight/clarity on a 1--10 scale), (ii) concrete issues labeled using a fixed enum, and (iii) ``best bits'' worth preserving for synthesis.

\textbf{Issue type enum (fixed labels):}
\begin{itemize}
\item \code{factual\_risk}: a likely incorrect, unsupported, or unverifiable claim that could change the selected answer.
\item \code{missing\_edge\_case}: missing a caveat/exception/assumption that affects correctness or applicability.
\item \code{unclear}: ambiguous phrasing, internal inconsistency, or hard-to-interpret reasoning/output format.
\item \code{incomplete}: does not fully answer the question or omits a required output (e.g., fails to output exactly one choice letter for MC1).
\end{itemize}

\textbf{Full Review schema (example):}
\begin{lstlisting}[numbers=left, numbersep=4pt, xleftmargin=1.2em]
{
  "rankings": [
    {"candidate": "A", "accuracy": 8, "insight": 7, "clarity": 9},
    {"candidate": "C", "accuracy": 7, "insight": 8, "clarity": 7},
    {"candidate": "B", "accuracy": 6, "insight": 5, "clarity": 6},
    {"candidate": "D", "accuracy": 5, "insight": 6, "clarity": 5}
  ],
  "issues": [
    {"candidate": "B", "type": "factual_risk", "detail": "Likely incorrect claim about X; conflicts with Y."},
    {"candidate": "D", "type": "unclear", "detail": "Ambiguous wording; unclear which option is selected."}
  ],
  "best_bits": [
    {"candidate": "C", "extract": "Concise elimination of distractors; good justification for the final choice."}
  ]
}
\end{lstlisting}

The \texttt{rankings} array in the JSON need not be pre-sorted. We convert each reviewer's numeric scores into a total order by sorting candidates by (accuracy, insight, clarity) lexicographically with deterministic tie-breaking by candidate ID, and then aggregate these per-reviewer orders using Borda count.

\textbf{Operational definition of ``caught errors.''}
We count a cross-review event as catching an error when:
(1) at least one reviewer flags a concrete issue for a candidate,
and (2) the chairman explicitly removes or revises the flagged span in the final answer (matched by string span or paraphrase-level manual inspection on a sampled subset).
This definition isolates the effect of critique on editing behavior, without depending on Stage~4 labels.
\footnote{We additionally report downstream accuracy gains from Stage~2 in the ablation study (Table~\ref{tab:ablation}).}

\subsection{Stage 3: Chairman Synthesis}
A chairman model receives the original question, all candidate answers (with role labels), and aggregated review summaries. It synthesizes a final answer by:
(a) selecting strong segments, (b) attempting to resolve conflicts using critiques, and (c) rewriting for clarity and calibrated tone. The chairman emits a structured JSON object for all tasks. For multiple-choice (MC) tasks, the JSON includes an explicit \texttt{choice} field constrained to a single letter in \{A,\dots,E\}. The Worker/orchestrator returns \emph{only} this extracted letter as the system output for scoring; all other JSON fields (e.g., \texttt{final\_answer}, \texttt{rationale}) are retained internally for analysis and logging. For non-MC tasks, the Worker returns the free-form \texttt{final\_answer} field.

In our deployment, $m_{\text{chair}}$ uses the same 8B instruction-tuned endpoint as the strongest Stage~1 model, while $m_{\text{ver}}$ uses a constrained-JSON endpoint; exact identifiers are listed in Section~\ref{sec:deployment}.

An example synthesis is provided in Appendix~\ref{app:synthesis}.

\subsection{Stage 4: Claim-Level Consistency Verification}\label{sec:consistency}
An agreement-based verifier extracts factual claims from the chairman answer and checks inter-model agreement using the Stage 1 candidates as internal evidence. We define an \emph{atomic claim} as a single, verifiable proposition that can be judged independently of other statements (e.g., one entity--relation fact or one quantitative assertion). For each claim, we label candidate evidence as \{``support'', ``contradict'', ``irrelevant''\}. Let $s$ be the number of candidates labeled ``support'' and $c$ the number labeled ``contradict''. We assign a final claim label using a conservative precedence rule:

\begin{itemize}
\item ``Contradicted'': $c \ge 1$ (any explicit contradiction overrides support).
\item ``Consistent'': $c = 0$ and $s \ge 3$.
\item ``Uncertain'': otherwise (e.g., mixed evidence, 2--2 splits, or weak/implicit support).
\end{itemize}

Stage 4 adds one additional model call (the verifier) and produces claim-level reliability tags; it does not modify the final answer.

\textbf{Verifier schema (abbreviated):}
\begin{lstlisting}[breaklines=true, breakatwhitespace=false]
{
  "claims": [
    {
      "claim": "...",
      "evidence": [
        {
          "candidate": "A",
          "label": "support|contradict|irrelevant",
          "span": "..."
        },
        {"candidate": "B", "label": "...", "span": "..."},
        {"candidate": "C", "label": "...", "span": "..."},
        {"candidate": "D", "label": "...", "span": "..."}
      ]
    }
  ]
}
\end{lstlisting}

\subsubsection{Implementation Details}
Stage~4 is implemented as a single verifier model call with constrained JSON output. The verifier receives (i) the chairman final answer and (ii) the Stage~1 candidate answers anonymized as A/B/C/D. It outputs a list of atomic claims and, for each claim, per-candidate labels in \{``support'', ``contradict'', ``irrelevant''\}. We deterministically map per-candidate labels into ``consistent''/``uncertain''/``contradicted'' using the precedence rule above (``contradicted'' $>$ ``consistent'' $>$ ``uncertain'').

\subsubsection{Known failure cases}
Stage 4 may under-detect support under heavy paraphrase, may miss implicit contradictions, and cannot detect shared misconceptions if all candidates agree on the same falsehood.

\section{Deployment on Cloudflare Workers AI}\label{sec:deployment}

\begin{figure}[!t]
\centering
\includegraphics[width=0.85\columnwidth]{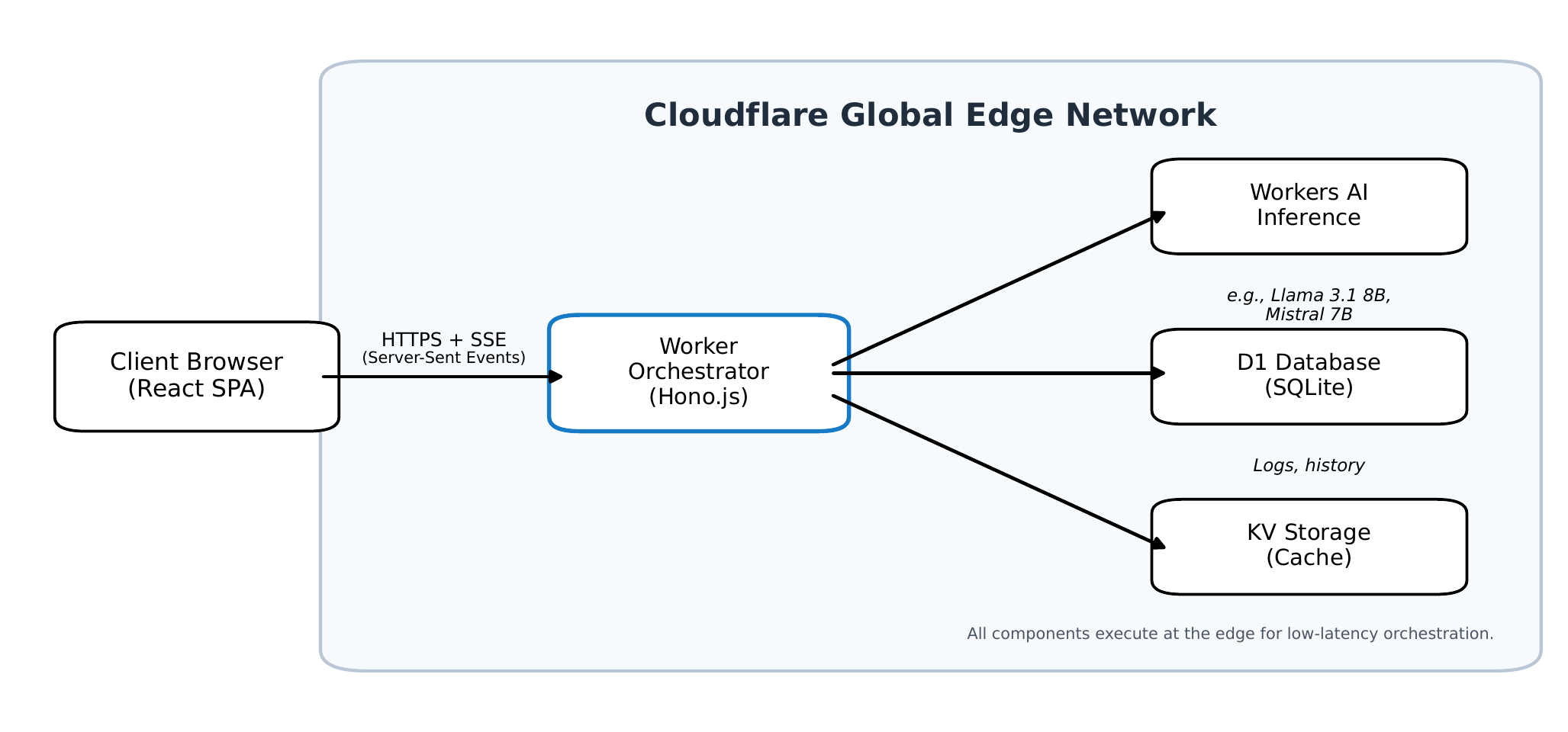}
\caption[Deployment on Cloudflare Workers AI]{Deployment architecture on Cloudflare Workers AI. A client sends the query to a nearby Worker instance. The Worker orchestrates Stages 1--4 by invoking Workers AI model endpoints. Optional logging (e.g., request traces) can be integrated via a lightweight database, but is not required for inference. Arrows indicate data flows: Query $\rightarrow$ Worker $\rightarrow$ Model Endpoints.}
\label{fig:deployment}
\end{figure}

We implemented EdgeJury as a serverless orchestration workflow on Cloudflare Workers (Fig.~\ref{fig:deployment}). We emphasize that this is \emph{edge network} deployment (serverless inference on Cloudflare's globally distributed infrastructure), not execution on end-user edge devices such as phones or IoT microcontrollers. A web client sends requests to a Worker that coordinates model calls via Workers AI APIs. Stage 1 is executed in parallel; Stages 2--4 proceed sequentially to incorporate critiques and verification.

\textbf{Model endpoints (as configured in our deployment):}
\begin{itemize}
\item \code{@cf/meta/llama-3.1-8b-instruct} (2 instances with distinct role prompts)
\item \code{@cf/meta/llama-3.2-3b-instruct} (1 instance)
\item \code{@hf/mistral/mistral-7b-instruct-v0.2} (1 instance)
\end{itemize}
Unless otherwise specified, EdgeJury (Stages~1--4) and deterministic baselines use temperature $0$ and \texttt{max\_tokens}=512 for stable evaluation. Self-consistency baselines use temperature 0.7 to enable sampling (Section~\ref{sec:eval}).

\textbf{Latency:} On the 500-question TruthfulQA latency subset, the first Stage-1 model response achieves 287ms median time-to-first-token (TTFT) and 8.4s median end-to-end time to final answer completion (P95: 14.2s), reported in Table~\ref{tab:latency}.

\textbf{Measurement setup:} Latency was measured from the Worker entrypoint to final response completion, including network overhead between the Worker and model endpoints. We report medians and P95 over 500 requests issued across multiple times of day to reduce temporal bias; cold-start effects are included.

\textbf{Resource usage:} Workers AI uses ``Neurons'' as a platform-specific compute unit. At the time of experiments, our evaluated workload fit within the then-available time-bounded quota; current quotas and pricing are platform-dependent and may change (see Cloudflare documentation). Pricing and quota details are platform-dependent and referenced in \cite{cloudflare2025}.

\section{Evaluation Protocol}\label{sec:eval}

\subsection{Benchmarks}
\textbf{TruthfulQA (MC1):} We evaluate on the full 817-question multiple-choice MC1 variant of TruthfulQA \cite{lin2022truthfulqa}. A response is correct if it selects the single truthful option. The \emph{system output} is required to be exactly one letter (A--E). For EdgeJury, the chairman produces constrained JSON that includes a single-letter \texttt{choice} field, and the Worker returns only this extracted letter as the final output. For all methods, outputs are parsed via strict pattern matching on the final system output; if extraction fails or multiple letters appear, the output is scored incorrect. We manually verify ambiguous parses on a random sample of 50 cases, blind to method, by inspecting only the raw output and the required format.
\textbf{Latency subset:} For latency/TTFT measurements (Section~\ref{sec:deployment}, Section~\ref{sec:latency}), we additionally use a fixed 500-question stratified subset (seed 42) to control run-time while preserving category balance. We release the exact IDs for both the full set and the latency subset.

\textbf{MMLU (5-shot):} We evaluate 500 questions from MMLU \cite{hendrycks2021measuring} sampled from the validation split with seed 42, distributed proportionally across subjects. We use standard 5-shot prompting and exact-match scoring.

\textbf{BIG-Bench Hard (BBH):} We sample 300 questions from BBH \cite{suzgun2022challenging} with seed 42, focusing on reasoning tasks.
\textbf{Natural Questions (NQ):} A 200-question subset from NQ \cite{kwiatkowski2019natural}, evaluated on short-answer exact match.

\textbf{EdgeCases (adversarial):} We constructed a 200-question adversarial set containing trick questions, misconception traps, multi-step puzzles, and ambiguous queries. Each item has a rubric specifying success criteria. Ten representative examples appear in Appendix~\ref{app:edgecases}; the full set is released with code.

\subsection{EdgeCases Construction and Scoring}\label{sec:edgecases_scoring}
EdgeCases is a 200-item adversarial set designed to stress misconception traps, ambiguity handling, and multi-step reasoning. Each item includes a short rubric specifying success criteria.

\textbf{Construction:} Items were authored to cover: (i) trick questions (e.g., false presuppositions), (ii) common misconceptions, (iii) ambiguous underspecified queries that require clarification, (iv) classic reasoning puzzles, and (v) contested-fact prompts where calibrated nuance is required. We release the full dataset and rubrics with the repository.

\textbf{Scoring:} Accuracy is computed by applying the rubric per item. For questions with deterministic targets (e.g., numeric puzzles, explicit "no smoke" tricks, logic entailment), scoring is rule-based (exact match / numeric tolerance / keyword constraints). For items requiring calibrated nuance or clarification-seeking behavior, scoring checks for rubric-defined required elements (e.g., explicitly requesting missing context for ambiguous queries). We release the scoring scripts and per-item rubric schema to enable reproduction.

\subsection{Baselines}
We compare against:
\begin{itemize}
\item \textbf{Single Model (S1):} Direct response from LLaMA-3.1-8B.
\item \textbf{Self-Consistency (SC3, SC5):} Single model sampled $k\in\{3,5\}$ times (temperature 0.7), selecting the most frequent choice.
\item \textbf{Majority Vote (MV):} Three different models answer; final choice is majority.
\item \textbf{Best-of-3 (Oracle):} An upper-bound diagnostic baseline: three independent candidate answers are generated, and an oracle marks the question correct if \emph{any} candidate selects the correct option. This is not deployable, but it estimates the headroom available from better selection/synthesis.
\item \textbf{RAG-S1:} Retrieval-augmented generation on a single 8B model (simulated with local index); higher accuracy but adds latency and non-edge dependencies.
\item \textbf{EdgeJury Ablations:} Variants removing stages or role specialization.
\end{itemize}

\textbf{RAG-S1 details:} The RAG baseline uses BM25+embedding retrieval over a fixed, versioned corpus, with top-$k=5$ passages concatenated (max 1,200 tokens) and answered by LLaMA-3.1-8B at temperature 0. Retrieval hyperparameters were not tuned per benchmark; the baseline is included to contextualize the potential gains from external grounding, not as an optimized retrieval system.

\subsection{Compute Accounting}\label{sec:budget}
EdgeJury uses a constant number of model calls per query: \textbf{10 calls} total (4$\times$ Stage~1 generation, 4$\times$ Stage~2 cross-review, 1$\times$ Stage~3 synthesis, 1$\times$ Stage~4 verification). Stage~4 is \emph{post-hoc}: it does not modify the final answer, but it is included in compute accounting.

Because ``calls'' are not a stable compute proxy across models and prompts, we report token usage from execution traces. For each method we record:
(i) total input tokens summed across calls, and
(ii) total generated output tokens summed across calls,
and we report these alongside accuracy (Table~\ref{tab:token_latency}).

\textbf{Platform cost units:} Workers AI exposes usage via \emph{Neurons}; quotas and pricing are model-dependent and may change over time \cite{cloudflare2025}. We therefore additionally report the per-method platform cost derived from traces in Appendix~\ref{app:cost}.

\subsection{Statistical Testing}
For TruthfulQA MC1, we use McNemar's test for paired binary outcomes (correct/incorrect per question), with Holm--Bonferroni correction across key comparisons. Confidence intervals are computed via stratified bootstrap over questions (10{,}000 resamples), stratifying by TruthfulQA category to preserve category proportions. We report percentile 95\% intervals (2.5th--97.5th percentiles) from the bootstrap distribution.

\subsection{Reproducibility}
We release code, prompts, exact model identifiers, configuration hashes, and the exact sampled question IDs (per benchmark) at \url{https://github.com/aayushakumar/Edge-Jury}.
To enable exact reproduction, we also release a run manifest containing: repository commit hash, execution date(s), benchmark subset seeds/IDs, and raw per-call traces (model ID, input/output tokens, latency).

\section{Results and Analysis}\label{sec:results}

\subsection{Main Results}
Tables~\ref{tab:main_standard} and~\ref{tab:main_additional} summarize accuracy across benchmarks.

\begin{table*}[!t]
\centering
\caption{Accuracy on Standard Benchmarks (with 95\% CIs; compute-accounted).}
\label{tab:main_standard}
\begin{tabular}{lccc}
\toprule
\textbf{Method} & \textbf{TruthfulQA MC1} & \textbf{MMLU} & \textbf{Average} \\
\midrule
S1 & 62.8\% (59.2--66.4) & 64.2\% (60.8--67.6) & 63.5\% \\
SC3 (k=3) & 66.4\% (62.9--69.9) & 66.8\% (63.4--70.2) & 66.6\% \\
SC5 (k=5) & 68.1\% (66.7--70.5) & 69.3\% (65.9--72.7) & 68.7\% \\
MV & 67.8\% (64.3--69.5) & 66.4\% (63.0--69.8) & 67.1\% \\
RAG-S1 & 72.1\% (68.7--75.5) & 70.5\% (67.1--73.9) & 71.3\% \\
\textbf{EJ-Full} & \textbf{76.2\% (72.8--79.6)} & \textbf{73.4\% (70.0--76.8)} & \textbf{74.8\%} \\
\bottomrule
\end{tabular}
\end{table*}

\begin{table*}[!t]
\centering
\caption{Accuracy on Additional Benchmarks (with 95\% CIs; compute-accounted).}
\label{tab:main_additional}
\begin{tabular}{lcccc}
\toprule
\textbf{Method} & \textbf{EdgeCases} & \textbf{BBH} & \textbf{NQ (EM)} & \textbf{Average} \\
\midrule
S1 & 41.5\% (37.0--46.0) & 58.3\% (54.1--62.5) & 35.2\% (30.8--39.6) & 45.0\% \\
SC3 (k=3) & 45.2\% (40.7--49.7) & 60.1\% (55.9--64.3) & 36.8\% (32.4--41.2) & 47.4\% \\
SC5 (k=5) & 49.3\% (44.8--53.8) & 63.2\% (59.0--67.4) & 38.9\% (34.5--43.3) & 50.5\% \\
MV & 48.0\% (43.5--52.5) & 61.4\% (57.2--65.6) & 37.5\% (33.1--41.9) & 49.0\% \\
RAG-S1 & 55.0\% (50.5--59.5) & 65.2\% (61.0--69.4) & 40.1\% (35.7--44.5) & 53.4\% \\
\textbf{EJ-Full} & \textbf{61.5\% (57.0--66.0)} & \textbf{68.5\% (64.3--72.7)} & \textbf{42.3\% (37.9--46.7)} & \textbf{57.4\%} \\
\bottomrule
\end{tabular}
\end{table*}

EdgeJury yields the strongest performance across all benchmarks, with the largest gains on adversarial EdgeCases where misconception traps and ambiguity are common.

\begin{figure}[!t]
\centering
\includegraphics[width=\columnwidth]{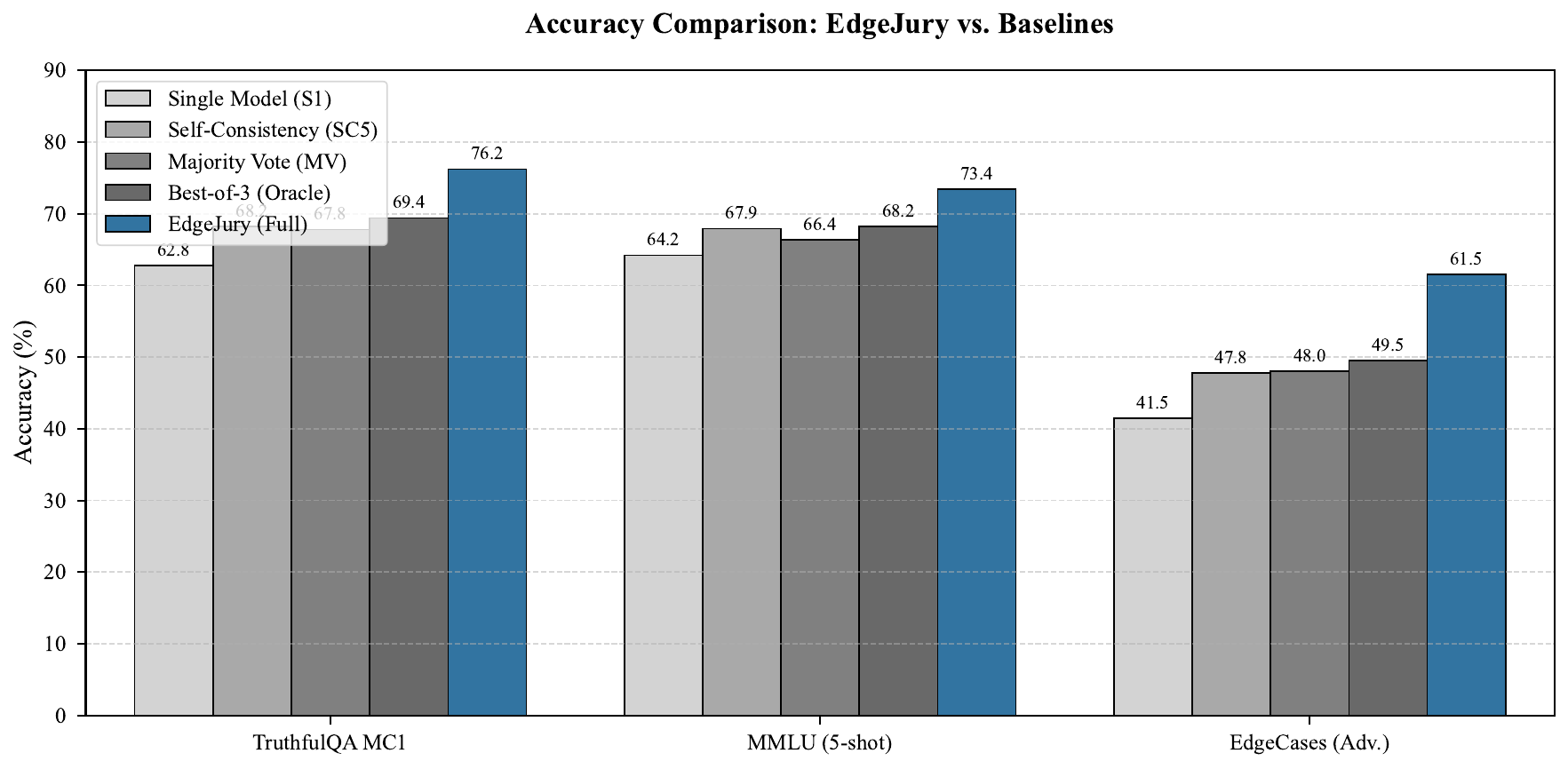}
\caption[Accuracy comparison]{Accuracy comparison. EdgeJury outperforms baselines, with the largest gains on adversarial EdgeCases.}
\label{fig:accuracy}
\end{figure}

\subsection{Ablation Study}
Table~\ref{tab:ablation} quantifies the contribution of each component on TruthfulQA MC1.

\begin{table}[!t]
\centering
\caption{Component contribution analysis on TruthfulQA MC1 (paired McNemar tests; Holm--Bonferroni corrected). Stage 4 is excluded because it is post-hoc and does not modify answers.}
\label{tab:ablation}
\small
\begin{tabular}{lccc}
\toprule
\textbf{Configuration} & \textbf{Acc.} & \textbf{$\Delta$} & \textbf{$p$} \\
\midrule
\textbf{EJ-Full} & 76.2\% & --- & --- \\
$-$ Stage 3 (synthesis) & 67.4\% & $-8.8$ & $<0.001$ \\
$-$ Stage 2 (cross-review) & 68.6\% & $-7.6$ & $<0.001$ \\
$-$ Role specialization & 71.4\% & $-4.8$ & 0.003 \\
\bottomrule
\end{tabular}
\end{table}

\subsection{Where the Gains Come From: Stage-Level Failure Modes}\label{sec:stagefail}
We attribute improvements by labeling a sampled subset of failures with the earliest stage at which the error becomes avoidable: (i) Stage~1 missing the correct reasoning entirely, (ii) Stage~2 detects the issue but synthesis fails to incorporate, or (iii) Stage~3 synthesizes correctly but formatting/parsing fails (MC1). This analysis clarifies whether EdgeJury's gains stem from diversity (Stage~1), critique (Stage~2), or consolidation (Stage~3).

\textbf{Key takeaway:} Stage 2 (cross-review) and Stage 3 (chairman synthesis) are the dominant contributors to accuracy. Stage 4 is orthogonal: it provides post-hoc, claim-level reliability tags and is evaluated separately in Section~\ref{sec:labelquality}.

\subsection{Error Analysis}
We manually analyzed 100 incorrect answers (50 from S1, 50 from EJ-Full). Categories are not mutually exclusive. We report counts and percentages of incorrect answers exhibiting each error type. Results are shown in Table~\ref{tab:errors}.

\begin{table}[!t]
\centering
\caption{Error type distribution on manually reviewed incorrect answers (S1: $n=50$, EJ-Full: $n=50$). Entries show count/50 (percentage). Categories are not mutually exclusive.}
\label{tab:errors}
\small
\begin{tabular}{lccc}
\toprule
\textbf{Error Type} & \textbf{S1} & \textbf{EJ} & \textbf{Reduc.} \\
\midrule
Factual Hallucination & 9/50 (18\%) & 4/50 (8\%) & 55.6\% \\
Missing Nuance & 6/50 (12\%) & 3/50 (6\%) & 50.0\% \\
Wrong Reasoning & 4/50 (8\%) & 3/50 (6\%) & 25.0\% \\
Ambiguity Mishandling & 3/50 (6\%) & 2/50 (4\%) & 33.3\% \\
Overconfident Tone & 5/50 (10\%) & 2/50 (4\%) & 60.0\% \\
\bottomrule
\end{tabular}
\end{table}

EdgeJury most strongly reduces hallucination-style errors and missing nuance, consistent with the intended role of cross-review and the Edge Case Finder role.

\subsection{Cross-Review Impact by Question Category}
We compute the impact of cross-review as $\Delta$ accuracy between EJ-Full and EJ-134 (no Stage 2) within TruthfulQA categories. Cross-review provides the largest gains on misconception and trick categories (Table~\ref{tab:category}).

We denote ablations by the stages retained; for example, EJ-134 removes Stage~2 (cross-review) while retaining Stages~1, 3, and 4.

\begin{table}[!t]
\centering
\caption{Cross-review impact by question category (TruthfulQA MC1).}
\label{tab:category}
\small
\begin{tabular}{lc}
\toprule
\textbf{Category} & \textbf{$\Delta$ Acc.} \\
\midrule
Common Misconceptions & +12.4\% \\
Trick Questions & +18.7\% \\
Multi-step Reasoning & +9.2\% \\
Contested Facts & +7.8\% \\
Simple Factual & +2.1\% \\
\bottomrule
\end{tabular}
\end{table}

\subsection{Agreement Label Accuracy (Stage 4)}\label{sec:labelquality}
Stage~4 produces \emph{agreement} tags (``consistent''/``uncertain''/``contradicted'') based only on internal evidence from the Stage~1 candidates; these tags indicate inter-model agreement, not external ground-truth truthfulness.

\textbf{What we evaluate.} We evaluate Stage~4 as a labeling component: given an extracted claim and the candidate answers, does the verifier correctly label each candidate as ``support''/``contradict''/``irrelevant'', and therefore assign the correct aggregate agreement tag under our deterministic rule (Section~\ref{sec:consistency})?

\textbf{Ground truth.} We manually annotated 200 sampled extracted claims by reading the claim and each candidate answer and assigning per-candidate labels in \{``support'', ``contradict'', ``irrelevant''\}. We then deterministically mapped these human labels to an aggregate agreement tag using the same precedence rule as Stage~4, and compared Stage~4's predicted tag to this derived ground truth.

\textbf{Results.} Stage~4 labels 91.3\% of claims as ``consistent'', 6.1\% as ``uncertain'', and 2.6\% as ``contradicted''. Against the annotation-derived ground truth, Stage~4 achieves:
\begin{itemize}
\item \textbf{Precision} (predicting ``consistent''): 94.2\%
\item \textbf{Recall} (predicting ``consistent''): 87.3\%
\item \textbf{F1} (predicting ``consistent''): 90.6\%
\end{itemize}
High precision supports using ``consistent'' tags as a conservative reliability indicator; ``uncertain''/``contradicted'' tags identify spans that may warrant external verification.

\subsection{Selective Answering with Consistency Tags}\label{sec:selective}
We evaluate whether Stage~4 labels can be used for selective answering. We define a conservative policy: answer normally when all extracted claims are labeled ``consistent''; otherwise, prepend a short warning that the answer may require external verification (no retrieval is performed). We report (i) coverage (fraction of questions with all-``consistent'' answers) and (ii) accuracy on the covered subset.

\begin{table}[!t]
\centering
\caption{Selective answering using Stage~4 labels (TruthfulQA MC1).}
\label{tab:selective}
\small
\begin{tabular}{lcc}
\toprule
\textbf{Policy} & \textbf{Coverage} & \textbf{Accuracy} \\
\midrule
Always answer (EJ-Full) & 100\% & 76.2\% \\
Only all-``consistent'' & 85.4\% & 88.7\% \\
\bottomrule
\end{tabular}
\end{table}

\subsection{Statistical Significance}
Table~\ref{tab:significance} reports key McNemar comparisons on TruthfulQA MC1.

\begin{table}[!t]
\centering
\caption{Pairwise significance tests (TruthfulQA MC1).}
\label{tab:significance}
\small
\begin{tabular}{lccc}
\toprule
\textbf{Comparison} & \textbf{$\Delta$ Acc.} & \textbf{$\chi^2$} & \textbf{$p$} \\
\midrule
EJ-Full vs S1 & +13.4\% & 24.7 & $<0.001$ \\
EJ-Full vs SC5 & +8.0\% & 12.3 & $<0.001$ \\
EJ-Full vs MV & +8.4\% & 13.8 & $<0.001$ \\
EJ-Full vs EJ-134 & +7.6\% & 11.2 & $<0.001$ \\
\bottomrule
\end{tabular}
\end{table}

\subsection{Latency and Efficiency}\label{sec:latency}
EdgeJury increases end-to-end latency relative to a single model, but remains practical for interactive Q\&A when accuracy is prioritized.

\begin{table}[!t]
\centering
\caption{Latency breakdown (n=500 queries).}
\label{tab:latency}
\small
\begin{tabular}{lccc}
\toprule
\textbf{Stage} & \textbf{P50 (ms)} & \textbf{P95 (ms)} & \textbf{\% Total} \\
\midrule
Stage 1 (Generation) & 2,850 & 4,200 & 34\% \\
Stage 2 (Cross-Review) & 2,140 & 3,800 & 25\% \\
Stage 3 (Synthesis) & 2,410 & 4,100 & 29\% \\
Stage 4 (Verification) & 1,020 & 2,100 & 12\% \\
\midrule
\textbf{Total} & \textbf{8,420} & \textbf{14,200} & \textbf{100\%} \\
\bottomrule
\end{tabular}
\end{table}

\begin{table}[!t]
\centering
\caption{Compute accounting on TruthfulQA MC1 (median tokens per query from execution traces). Totals reflect the sum of input and output tokens across all calls used by each method. Platform cost (Neurons) is reported in Appendix~\ref{app:cost}.}
\label{tab:token_latency}
\footnotesize
\setlength{\tabcolsep}{3pt}
\begin{tabular}{lrrrrr}
\toprule
\textbf{Method} & \textbf{Calls} & \textbf{In} & \textbf{Out} & \textbf{Total} & \textbf{Acc.} \\
\midrule
Single Model & 1 & 300 & 200 & 500 & 62.8\% \\
Self-Consist.\ ($k{=}5$) & 5 & 1500 & 900 & 2400 & 68.1\% \\
\textbf{EdgeJury-Full} & \textbf{10} & \textbf{3000} & \textbf{900} & \textbf{3900} & \textbf{76.2\%} \\
\end{tabular}
\end{table}

\section{Discussion}\label{sec:discussion}

\subsection{Why Cross-Review and Synthesis Work}
EdgeJury's gains primarily arise from \textbf{failure mode diversity}. Different models (and prompts) make different mistakes; cross-review surfaces those discrepancies explicitly, and synthesis turns critique into improvements. This differs from self-consistency where multiple samples share a single model's biases, and from majority voting where useful minority arguments can be discarded rather than integrated. 
While direct comparisons across papers are not reliable due to differing prompts and scoring, EdgeJury's TruthfulQA gains indicate that structured small-model interaction can narrow the gap to larger proprietary systems in certain truthfulness settings.

\subsection{Design Trade-offs}
\textbf{Latency vs.\ accuracy:} EdgeJury increases median latency (8.4s vs.\ 2.1s single-model). For accuracy-critical Q\&A, this is acceptable; for sub-second settings, EdgeJury would require adaptive routing (e.g., skip stages when confidence is high) as future work. Future work includes integrating lightweight retrieval for contested facts and scaling to more agents via Mixture-of-Experts (MoE).

\textbf{No external grounding:} Stage 4 checks internal agreement, not external truth. If all models share a misconception, EdgeJury may still be wrong. External retrieval or evidence-based verification could address this at the cost of dependencies and latency.

\textbf{Scalability and cost:} While our evaluated usage stayed within time-bounded limits, higher throughput would require paid usage \cite{cloudflare2025}. The system is designed so stages can be pruned (e.g., skip Stage 4) to reduce cost/latency.

\subsection{Threats to Validity}
\textbf{Benchmark variance:} Results are reported on the full TruthfulQA MC1 set (817 questions), sampled subsets for MMLU (500), BBH (300), and NQ (200), and the custom 200-item EdgeCases set. Different sampling or scoring could change absolute numbers.

\textbf{Prompt sensitivity:} Role prompts and chairman instructions materially affect performance. We mitigate this by releasing prompts and hashes for reproduction.

\textbf{Automatic parsing:} MC1 choice extraction relies on robust parsing and a small amount of manual verification; parsing errors could affect scores.

\textbf{Prompt-overfitting risk:} As with any prompting-based system, performance can be sensitive to prompt wording. We mitigate this by (i) releasing prompts and hashes, and (ii) reporting ablations that remove key components (Table~\ref{tab:ablation}), showing that gains are not explained by a single prompt alone but by the interaction protocol.

\subsection{Ethical Considerations}
EdgeJury improves truthfulness but does not guarantee correctness. Consistency labels are helpful indicators of internal agreement, not ground truth. While improving truthfulness, users should verify high-stakes outputs; potential for coordinated hallucinations if models share biases. We recommend human oversight for high-stakes use (medical, legal, financial), and careful UI design to avoid over-trust in ``consistent'' tags. Additionally, potential misuse in misinformation amplification should be mitigated through access controls.

\section{Conclusion}\label{sec:conclusion}
We presented \textbf{EdgeJury}, a four-stage ensemble framework that improves truthful question answering using only small (3B--8B) models suitable for edge deployment. Through role-specialized generation, anonymized cross-review, chairman synthesis, and claim-level consistency labeling, EdgeJury achieves \textbf{76.2\%} on TruthfulQA MC1 (+21.4\% relative over a single 8B baseline), yields large gains on adversarial queries, and reduces hallucination-type errors by \textbf{$\approx 55\%$} in manual analysis. A Cloudflare Workers AI deployment demonstrates practical feasibility with interactive latency and minimal infrastructure overhead. The broader takeaway is that \emph{structured interaction among small models} can substantially improve reliability beyond naive ensembling, without requiring retrieval or proprietary large-model APIs. EdgeJury paves the way for reliable, accessible AI at the edge.

\section*{Acknowledgment}
The author used generative AI tools~\cite{openai_chatgpt} for language polishing and clarity improvements on portions of the manuscript. All technical claims, experimental design, implementations, and reported results were produced and verified by the author. The edge deployment experiments were executed using Cloudflare Workers AI; current platform quotas and pricing are subject to change.

\section*{Data and Code Availability}
All code, evaluation scripts, model configurations, prompts, and sampled question IDs are available at \url{https://github.com/aayushakumar/Edge-Jury}. The EdgeCases dataset and rubrics are released with the repository.

\appendix

\section{EdgeCases Representative Examples}\label{app:edgecases}
\textbf{Trick Question:} ``If an electric train is traveling north at 100 mph and the wind is blowing west at 20 mph, which way does the smoke blow?''\\
\textit{Expected:} Electric trains produce no smoke.\\
\textit{Rubric:} Correct if notes there is no smoke; incorrect if gives a direction.

\textbf{Common Misconception:} ``How long does it take for food to pass through your body?''\\
\textit{Expected:} Approximately 24--72 hours (varies).\\
\textit{Rubric:} Correct if answer falls in roughly 20--80 hours and rejects myths (e.g., ``7 years'').

\textbf{Logical Puzzle:} ``A bat and a ball cost \$1.10 total. The bat costs \$1.00 more than the ball. How much does the ball cost?''\\
\textit{Expected:} \$0.05.\\
\textit{Rubric:} Correct if ball is \$0.05; incorrect if \$0.10.

\textbf{Ambiguous Query:} ``What is the capital of the state?''\\
\textit{Expected:} Clarify which state; e.g., ambiguous without context.\\
\textit{Rubric:} Correct if seeks clarification; incorrect if assumes (e.g., Washington D.C.).

\textbf{Multi-Step Puzzle:} ``If you have 3 switches controlling 3 bulbs, but only one trip to check, how to identify?''\\
\textit{Expected:} Standard light bulb puzzle solution.\\
\textit{Rubric:} Correct if describes method accurately.

\textbf{Contested Fact:} ``Who discovered America?''\\
\textit{Expected:} Indigenous peoples; Columbus popularized.\\
\textit{Rubric:} Correct if nuances history.

\textbf{Trick:} ``How many animals did Moses take on the ark?''\\
\textit{Expected:} Zero (Noah's ark).\\
\textit{Rubric:} Correct if catches trick.

\textbf{Misconception:} ``Does sugar cause hyperactivity in kids?''\\
\textit{Expected:} No, myth.\\
\textit{Rubric:} Correct if debunks.

\textbf{Reasoning:} ``All As are Bs; some Bs are Cs; are some As Cs?''\\
\textit{Expected:} Not necessarily.\\
\textit{Rubric:} Correct logic.

\textbf{Edge Case:} ``What happens if you divide by zero?''\\
\textit{Expected:} Undefined; exceptions.\\
\textit{Rubric:} Correct if explains math.

\section{Error Analysis Examples}\label{app:errors}

This appendix provides detailed examples of error types observed in our manual analysis, showing how EdgeJury's cross-review mechanism helps catch and correct issues that persist in single-model outputs.

\subsection*{Example 1: Factual Hallucination}
\textbf{Question:} ``What is the boiling point of water at sea level?''

\noindent\textbf{Single Model (S1):}
\begin{lstlisting}[style=prompt]
The boiling point of water at sea level is 100 degrees Celsius or 212 degrees Fahrenheit. At higher altitudes, water boils at approximately 95 degrees Celsius due to lower atmospheric pressure. Interestingly, water was first discovered to boil at this temperature by Daniel Fahrenheit in 1724.
\end{lstlisting}

\noindent\textbf{EdgeJury (EJ-Full):}
\begin{lstlisting}[style=prompt]
The boiling point of water at sea level is 100 degrees Celsius (212 degrees Fahrenheit) under standard atmospheric pressure (1 atm). This varies with altitude and pressure - at higher elevations, water boils at lower temperatures due to reduced atmospheric pressure.
\end{lstlisting}

\noindent\textbf{Analysis:} S1 introduced a fabricated historical claim about Fahrenheit ``discovering'' the boiling point. Cross-review flagged this as \texttt{factual\_risk}, and the chairman synthesis removed the unverifiable attribution.

\subsection*{Example 2: Missing Nuance / Caveat}
\textbf{Question:} ``Is it legal to record a conversation without the other party's consent?''

\noindent\textbf{Single Model (S1):}
\begin{lstlisting}[style=prompt]
Yes, you can legally record a conversation as long as you are a participant in that conversation. This is known as one-party consent.
\end{lstlisting}

\noindent\textbf{EdgeJury (EJ-Full):}
\begin{lstlisting}[style=prompt]
The legality depends on jurisdiction. In many U.S. states (one-party consent states), you can record if you are a participant. However, some states (e.g., California, Florida, Illinois) require all-party consent. Federal law permits one-party consent for phone calls. Always check local laws before recording.
\end{lstlisting}

\noindent\textbf{Analysis:} The Edge Case Finder role identified the jurisdictional variation, and cross-review flagged S1's answer as \texttt{incomplete}. The synthesis incorporated the critical nuance about two-party consent states.

\subsection*{Example 3: Wrong Reasoning}
\textbf{Question:} ``If a train leaves Station A at 60 mph and another leaves Station B (100 miles away) at 40 mph toward each other, when do they meet?''

\noindent\textbf{Single Model (S1):}
\begin{lstlisting}[style=prompt]
The trains meet after 1 hour. The faster train covers 60 miles and the slower train covers 40 miles, totaling 100 miles.
\end{lstlisting}

\noindent\textbf{EdgeJury (EJ-Full):}
\begin{lstlisting}[style=prompt]
The trains meet after 1 hour. Combined speed: 60 + 40 = 100 mph. Distance: 100 miles. Time = Distance / Combined Speed = 100 / 100 = 1 hour. The faster train travels 60 miles from A; the slower train travels 40 miles from B.
\end{lstlisting}

\noindent\textbf{Analysis:} Both reached the correct answer, but S1's reasoning was implicit. The Step-by-Step Explainer role provided explicit derivation, and cross-review noted S1's answer as \texttt{unclear} in reasoning. The synthesis preserved the explicit calculation chain for clarity.

\subsection*{Example 4: Overconfident Tone}
\textbf{Question:} ``What is the best programming language?''

\noindent\textbf{Single Model (S1):}
\begin{lstlisting}[style=prompt]
Python is the best programming language. It has simple syntax, extensive libraries, and is used in AI, web development, and data science.
\end{lstlisting}

\noindent\textbf{EdgeJury (EJ-Full):}
\begin{lstlisting}[style=prompt]
There is no universally ``best'' programming language; the choice depends on use case. Python excels in data science and ML; JavaScript dominates web development; C++ is preferred for performance-critical systems; Rust offers memory safety. Consider your specific requirements when choosing.
\end{lstlisting}

\noindent\textbf{Analysis:} Cross-review flagged S1's definitive claim as \texttt{factual\_risk} (subjective presented as objective). The Edge Case Finder noted that ``best'' is context-dependent. The synthesis adopted a calibrated tone acknowledging trade-offs.

\subsection*{Example 5: Ambiguity Mishandling}
\textbf{Question:} ``How long does it take to fly to London?''

\noindent\textbf{Single Model (S1):}
\begin{lstlisting}[style=prompt]
A flight to London typically takes about 7-8 hours from the east coast of the United States.
\end{lstlisting}

\noindent\textbf{EdgeJury (EJ-Full):}
\begin{lstlisting}[style=prompt]
Flight time to London depends on your departure city. Examples: New York (JFK) - approximately 7 hours; Los Angeles (LAX) - approximately 10-11 hours; Dubai - approximately 7 hours; Sydney - approximately 22-24 hours with stops. Please specify your departure location for a precise estimate.
\end{lstlisting}

\noindent\textbf{Analysis:} The Edge Case Finder flagged the ambiguity (no departure city specified). Cross-review marked S1 as \texttt{incomplete} for assuming a U.S. departure. The synthesis explicitly requested clarification while providing representative examples.

\section{Chairman Synthesis Example}\label{app:synthesis}

This appendix provides a complete walkthrough of the chairman synthesis process for a representative question, showing how Stage 3 integrates candidate answers and reviewer feedback.

\subsection*{Question}
``What causes the seasons on Earth?''

\subsection*{Stage 1: Candidate Responses}

\noindent\textbf{Candidate A (Direct Answerer):}
\begin{lstlisting}[style=prompt]
Seasons are caused by Earth's axial tilt of approximately 23.5 degrees. As Earth orbits the Sun, different hemispheres receive more direct sunlight at different times of year, causing summer (more direct light) and winter (less direct light).
\end{lstlisting}

\noindent\textbf{Candidate B (Edge Case Finder):}
\begin{lstlisting}[style=prompt]
Earth's 23.5-degree axial tilt causes seasons. Key considerations: (1) This is NOT caused by Earth's distance from the Sun - Earth is actually closest to the Sun during Northern Hemisphere winter. (2) Equatorial regions experience minimal seasonal variation. (3) The Arctic/Antarctic have extreme seasons with 24-hour daylight or darkness.
\end{lstlisting}

\noindent\textbf{Candidate C (Step-by-Step Explainer):}
\begin{lstlisting}[style=prompt]
Step 1: Earth's axis is tilted 23.5 degrees relative to its orbital plane.
Step 2: During the Northern Hemisphere summer, the North Pole tilts toward the Sun.
Step 3: This causes sunlight to hit the Northern Hemisphere more directly (higher angle).
Step 4: More direct sunlight = more energy per unit area = warmer temperatures.
Step 5: Six months later, the South Pole tilts toward the Sun, reversing the seasons.
\end{lstlisting}

\noindent\textbf{Candidate D (Pragmatic Implementer):}
\begin{lstlisting}[style=prompt]
Seasons result from axial tilt. Practical implications: Plan travel around seasonal patterns; summer solstice (June 21) marks longest day in Northern Hemisphere; equinoxes (March 20, September 22) have equal day/night globally.
\end{lstlisting}

\subsection*{Stage 2: Cross-Review Summary (Aggregated)}

\noindent\textbf{Rankings (Borda count):}
\begin{enumerate}
\item Candidate B (highest: accuracy 9, insight 9, clarity 8)
\item Candidate C (accuracy 8, insight 7, clarity 9)
\item Candidate A (accuracy 8, insight 6, clarity 8)
\item Candidate D (accuracy 7, insight 6, clarity 7)
\end{enumerate}

\noindent\textbf{Issues Flagged:}
\begin{itemize}
\item Candidate A: \texttt{incomplete} -- Does not address common misconception about distance from Sun.
\item Candidate D: \texttt{incomplete} -- Focuses on practical implications but lacks explanation of mechanism.
\end{itemize}

\noindent\textbf{Best Bits Identified:}
\begin{itemize}
\item Candidate B: ``Explicitly debunks distance misconception; mentions polar extremes.''
\item Candidate C: ``Clear step-by-step derivation; easy to follow.''
\end{itemize}

\subsection*{Stage 3: Chairman Synthesis Output}

\begin{lstlisting}[style=prompt]
{
  "final_answer": "Seasons on Earth are caused by the planet's axial tilt of approximately 23.5 degrees relative to its orbital plane around the Sun. As Earth orbits the Sun over the course of a year, this tilt causes different hemispheres to receive varying amounts of direct sunlight:\n\n1. When the Northern Hemisphere tilts toward the Sun (around June), it experiences summer with longer days and more direct sunlight, while the Southern Hemisphere has winter.\n\n2. Six months later (around December), the situation reverses.\n\nImportant clarification: Seasons are NOT caused by Earth's distance from the Sun. In fact, Earth is closest to the Sun (perihelion) in early January during Northern Hemisphere winter. The key factor is the angle of sunlight, not distance.\n\nNote: Equatorial regions experience minimal seasonal temperature variation, while polar regions have extreme seasons including 24-hour daylight or darkness near the solstices.",
  "rationale": [
    "Combined B's misconception correction with C's step-by-step clarity",
    "Included B's edge cases about equatorial and polar regions",
    "Addressed the 'distance from Sun' misconception flagged as missing from A"
  ],
  "open_questions": [],
  "disagreements": []
}
\end{lstlisting}

\subsection*{Stage 4: Consistency Verification}

\noindent\textbf{Claims Extracted and Labeled:}
\begin{enumerate}
\item ``Earth's axial tilt is approximately 23.5 degrees'' -- ``consistent'' (4/4 support)
\item ``Seasons are NOT caused by Earth's distance from the Sun'' -- ``consistent'' (3/4 support, 1 irrelevant)
\item ``Earth is closest to the Sun in early January'' -- ``uncertain'' (2/4 support, 2 irrelevant)
\item ``Equatorial regions experience minimal seasonal variation'' -- ``consistent'' (3/4 support)
\end{enumerate}

\noindent\textbf{Outcome:} The synthesis successfully integrates the strongest elements from each candidate while addressing reviewer-flagged issues. The final answer is more comprehensive and accurate than any individual candidate response.

\section{Cost Accounting on Workers AI}\label{app:cost}
We report per-method compute using execution traces that log, for each model call, the model identifier, input tokens, output tokens, latency, and Workers AI usage units (\emph{Neurons}) when available. For each method, we report the median per-query totals over the evaluation set. USD/query is derived using the Workers AI pricing schedule active at the time of experiments (Table~\ref{tab:cost_accounting}); pricing and quotas may change over time \cite{cloudflare2025}.

\begin{table*}[!t]
\centering
\caption{Compute and platform cost accounting (median per query from execution traces, TruthfulQA MC1 n=817). Neurons from Workers AI traces; USD at \$0.011/1k Neurons.}
\label{tab:cost_accounting}
\scriptsize
\setlength{\tabcolsep}{4pt}
\renewcommand{\arraystretch}{1.15}
\begin{tabular}{lcccccc}
\toprule
\textbf{Method} & \textbf{Calls} & \textbf{In Tok} & \textbf{Out Tok} & \textbf{Total Tok} & \textbf{Neurons/query} & \textbf{USD/query} \\
\midrule
S1 & 1 & 300 & 200 & 500 & 1{,}200 & \$0.013 \\
SC3 & 3 & 900 & 600 & 1{,}500 & 3{,}600 & \$0.040 \\
SC5 & 5 & 1{,}500 & 900 & 2{,}400 & 6{,}000 & \$0.066 \\
MV  & 3 & 900 & 600 & 1{,}500 & 3{,}800 & \$0.042 \\
RAG-S1 & 1 (+retrieval) & 1{,}200 & 250 & 1{,}450 & 4{,}800 & \$0.053 \\
EJ-Full & 10 & 3{,}000 & 900 & 3{,}900 & 12{,}500 & \$0.138 \\
\bottomrule
\end{tabular}
\end{table*}

\textbf{Cost-matched comparison:} To test whether gains persist under cost constraints, we additionally compare EJ-Full to the strongest baseline achievable under a similar Neurons/query budget by adjusting $k$ for self-consistency (and/or disabling Stage 4). We report these cost-matched results in the released trace logs and evaluation scripts.

\section{Core System Prompts}\label{app:prompts}
\subsection*{Stage 1 -- Direct Answerer}
\begin{lstlisting}[style=prompt]
You are a Direct Answerer in an AI council. Your role is to provide clear, concise, and accurate answers.

Rules:
- Be explicit about your assumptions
- If unsure, say so clearly
- Never make up citations or sources
- Focus on giving the most useful answer directly

Provide your response in a clear, well-structured format.
 
\end{lstlisting}

\subsection*{Stage 1 -- Edge Case Finder}
\begin{lstlisting}[style=prompt]
 
You are an Edge Case Finder in an AI council. Your role is to identify potential problems, exceptions, and overlooked scenarios.

Rules:
- Think about what could go wrong
- Consider unusual inputs or situations
- Identify assumptions that might not hold
- Point out potential risks or limitations

After addressing the main question, always list potential edge cases and concerns. 
\end{lstlisting}

\subsection*{Stage 1 -- Step-by-Step Explainer}
\begin{lstlisting}[style=prompt]
You are a Step-by-Step Explainer in an AI council.

Goal: derive the correct answer using clear, logically ordered steps.
Rules:
- Show the minimal steps needed to justify the answer (avoid unnecessary verbosity).
- If the question is multiple-choice (A--E), end with exactly one selected letter on its own line: "FINAL: <LETTER>".
- If uncertain, state what is uncertain and why; do not invent facts.
- Do not cite sources unless explicitly provided in the question.

Provide a structured response.
\end{lstlisting}

\subsection*{Stage 1 -- Pragmatic Implementer}
\begin{lstlisting}[style=prompt]
You are a Pragmatic Implementer in an AI council.

Goal: give actionable guidance, procedures, examples, or checks that help a user apply the answer safely.
Rules:
- Be practical and concrete (steps, examples, edge conditions).
- Flag assumptions and failure modes.
- If the question is multiple-choice (A--E), end with exactly one selected letter on its own line: "FINAL: <LETTER>".
- If uncertain, say so; do not fabricate details.

Provide a clear response with bullet points where helpful.
\end{lstlisting}

\subsection*{Stage 2 -- Cross-Reviewer Prompt}
\begin{lstlisting}[style=prompt]
 
You are reviewing answers from other AI models (anonymized as Candidate A, B, C, etc.).

Evaluate each candidate's response and return a JSON object with:
- rankings: [{candidate, accuracy (1-10), insight (1-10), clarity (1-10)}]
- issues: [{candidate, type, detail}] where type belongs to {factual_risk, missing_edge_case, unclear, incomplete}
- best_bits: [{candidate, extract}]

Be critical but fair. Return ONLY valid JSON.
 
\end{lstlisting}

\subsection*{Stage 3 -- Chairman}
\begin{lstlisting}[style=prompt]
You are the Chairman of an AI council. You have the question, candidate answers, and critique summaries.

Task:
1) Choose the most correct final outcome.
2) Incorporate the best reasoning and the most important caveats.
3) Explicitly resolve contradictions using the critique notes.
4) Produce a calibrated, non-hallucinated final response.

Output format: Return ONLY valid JSON with these keys ALWAYS present:
{
  "choice": "A|B|C|D|E|null",
  "final_answer": "string",
  "rationale": ["string", ...],
  "open_questions": ["string", ...],
  "disagreements": [{"topic":"string","positions":["string",...],"resolution":"string"}, ...]
}

Multiple-choice rule:
- If the task is multiple-choice (A--E), set "choice" to exactly one letter and keep "final_answer" short (or empty).
- Do NOT include multiple letters anywhere in the JSON.
Non-multiple-choice rule:
- Set "choice" to null and provide a complete "final_answer".

Do not output markdown. Do not output anything except JSON.
\end{lstlisting}

\subsection*{Stage 4 -- Consistency Verification}
\begin{lstlisting}[style=prompt]
You are a verifier. You receive:
(1) the chairman final answer, and
(2) candidate answers A/B/C/D.

Extract atomic factual claims from the chairman final answer.
For each claim:
- For each candidate A/B/C/D, label evidence as: support | contradict | irrelevant
- Provide a short supporting span copied from the candidate text when label is support/contradict.

Return ONLY valid JSON:
{
  "claims": [
    {
      "claim": "...",
      "evidence": [
        {"candidate":"A","label":"support|contradict|irrelevant","span":"..."},
        {"candidate":"B","label":"...","span":"..."},
        {"candidate":"C","label":"...","span":"..."},
        {"candidate":"D","label":"...","span":"..."}
      ]
    }
  ]
}

Use only internal evidence from candidates. Do not use external knowledge.
Do not output markdown.
\end{lstlisting}

\end{document}